# Wisdom of the AI Crowd (AI-CROWD) for Ground Truth Approximation in Content Analysis: A Research Protocol & Validation Using Eleven Large Language Models

Luis de-Marcos[1] (*), Manuel Goyanes[2], Adrián Domínguez-Díaz[1]

[1] Computer Science Department, Polytechnic School, University of Alcalá. Madrid, Spain.
[2] MITCOM Research Group, Department of Communication, Carlos III University of Madrid, Spain

Corresponding author: Luis de-Marcos (luis.demarcos@uah.es)

**Abstract:** Large-scale content analysis is increasingly limited by the absence of observable ground truth or gold-standard labels, as creating such benchmarks through extensive human coding becomes impractical for massive datasets due to high time, cost, and consistency challenges. To overcome this barrier, we introduce the AI-CROWD protocol, which approximates ground truth by leveraging the collective outputs of an ensemble of large language models (LLMs). Rather than asserting that the resulting labels are true ground truth, the protocol generates a consensus-based approximation derived from convergent and divergent inferences across multiple models. By aggregating outputs via majority voting and interrogating agreement/disagreement patterns with diagnostic metrics, AI-CROWD identifies high-confidence classifications while flagging potential ambiguity or model-specific biases.

**Highlights**

- We validate the protocol on four standard benchmarks (AG News, IMDb, DBpedia-14, SciCite) using 11 diverse LLMs in zero-shot mode.

- Majority-vote consensus closely approximates human ground truth, achieving macro-F1 scores up to 0.987 (DBpedia-14) and often rivaling or surpassing top individual models.

- Diagnostic metrics (Krippendorff's alpha, annotator alignment, skill-weighted task uncertainty) reveal strong reliability on structured tasks and highlight interpretive challenges on citation intent classification, supporting transparent and reflexive use in large-scale content analysis.

# Introduction

Large-scale content analysis has become a cornerstone of research in communication, computational social science, and related fields, driven by the growing availability of massive textual datasets such as news archives, social media streams, scientific publications, and policy documents (Budak et al., 2021; Joo & Steinert-Threlkeld, 2022; Goyanes & Piñeiro-Naval, 2024). While these data sources offer unprecedented opportunities for empirical inquiry, they also raise a fundamental methodological problem: in many large-scale settings, ground truth is either unavailable, undefined, or impossible to establish with sufficient certainty (Krippendorff, 2018; Araujo et al., 2020). When datasets reach a certain scale, constructing authoritative benchmark labels through extensive human annotation becomes impractical due to cost, time, and workload, leaving researchers without a clear benchmark against which inferences can be evaluated (Krippendorff, 2008; Lind et al., 2017).

Recent advances in large language models (LLMs) have substantially expanded the capacity to generate reliable automated content annotations at scale, enabling forms of inference that would otherwise be infeasible. Prior research has shown that LLM-generated classifications often align closely with human annotations across a wide range of tasks (Domínguez-Díaz et al., 2025). However, this literature does not fully address the core methodological challenge faced in many real-world applications: how to approximate ground truth itself when no external standard exists. In contexts characterized by data abundance, the central issue is not whether individual annotators (human or artificial) are reliable, but how researchers can construct approximations to ground truth through which inferences can be evaluated.

The research protocol proposed in this article is motivated by this gap. The AI-CROWD protocol approaches large-scale inference as a collective process based on the aggregation of multiple independent LLMs. Drawing on principles from crowdsourcing, the protocol treats LLMs as an ensemble of annotators whose convergent and divergent outputs provide informative signals of the quality of the inferences performed (Dumitrache, 2015; Zhang et al., 2016). Importantly, agreement across models is not assumed to imply validity; instead, agreement is interpreted as an indicator of relative confidence in the inferences performed, while disagreement is treated as a diagnostic signal revealing task difficulty, bias or coding errors (Dumitrache et al., 2018; Li et al., 2018; Nishio et al., 2025). In general, the AI-CROWD protocol does not seek to replace human judgment or to assert that aggregated AI outputs constitute truth. Rather, it provides a transparent and replicable method for approximating ground truth when external standards are unavailable. By combining ensemble-based aggregation with post-hoc diagnostic metrics, the protocol allows researchers to assess where collective AI inference converges, diverge, and how much confidence can reasonably be assigned to inferences.

Within this framework, approximating ground truth constitutes one important application of the proposed protocol, particularly in settings where no authoritative gold standard exists and large-scale human annotation is infeasible. However, this is not the sole use case of AI-CROWD. The protocol can also be employed as a strategy for enhancing annotation reliability in content analysis more broadly, even when the explicit goal is not to construct ground truth. In many applied settings, content coding relies typically on paired annotators, such as a single human coder paired with an automated system or two independent automated models or human annotators. While efficient, such approaches remain vulnerable to errors. By contrast, aggregating inferences across multiple independent LLMs enables the production of annotations that are systematically more robust than those generated by single human–AI pairings or isolated AI annotators. This may be an additional application of the AI-CROWD.

**Method details**

The Wisdom of the AI Crowd (AI-CROWD) protocol provides a scalable framework for approximating ground truth labels in large-scale content analysis tasks where human annotation is infeasible due to dataset size or resource constraints. By treating an ensemble of LLMs as a diverse "crowd" of annotators, the protocol generates a probabilistic consensus through aggregation, identifying areas of high agreement as reliable proxies for ground truth while flagging divergences for potential bias or uncertainty. This approach is particularly valuable in computational social science, where massive unstructured datasets (e.g., social media posts, news articles, or scientific abstracts) often lack gold standards. The protocol draws on crowdsourcing principles but adapts them to AI annotators, emphasizing transparency, diversity in the ensemble, and post-hoc diagnostics to ensure robustness. It consists of four sequential steps: dataset preparation, model-based coding with initial reliability analysis, consensus building via aggregation, and post-hoc analysis of the AI crowd's outputs. Below, each step is described in detail, including justifications for key decisions, required tools/parameters, and outputs. Implementation was conducted in Python (version 3.12), with full reproducibility via the accompanying code repository. Users applying the protocol to custom data should include also ethical considerations, such as data privacy and bias auditing, especially for sensitive domains.

Step 1: Dataset preparation

The foundation of the protocol is ensuring the input data is suitable for LLM annotation and aggregation. For optimal results, datasets should have clearly defined classification tasks with a closed set of mutually exclusive labels, as ambiguous or open-ended tasks can amplify inter-model divergences and reduce consensus reliability. This clarity minimizes prompt sensitivity and enhances the ensemble's ability to converge on meaningful approximations. To achieve this, we suggest developing a concise codebook or label schema that includes: (1) explicit definitions for each category with examples of prototypical instances, (2) boundary cases to illustrate distinctions, and (3) handling rules for edge cases (e.g., multi-label scenarios or incomplete text). This coding book serves as the basis for prompts in Step 2 and helps mitigate biases inherent in LLM pre-training.

If applying the protocol to custom data (e.g., user-collected corpora like social media archives or document repositories), preprocess the raw text to remove noise: normalize formatting (e.g., strip HTML tags, handle encoding issues), truncate overly long instances to fit LLM context windows (typically 4,000–128,000 tokens, depending on the model), and ensure balanced representation across potential classes if known. Use libraries like pandas (version 2.2 or later) for data manipulation and NLTK or spaCy for basic cleaning. For very large datasets (>100,000 instances), consider stratified sampling to create a manageable subset while preserving class distributions—justify the sample size based on computational resources and desired statistical precision (e.g., for metric confidence intervals).

For the methodological validation of the protocol, we used four standard benchmark datasets (AG News, IMDb, DBpedia-14, SciCite) to demonstrate applicability across diverse task types (thematic, sentiment, ontological, intent-based), knowledge requirements, and class counts (2–14). These were chosen for their established ground truth labels, enabling direct validation in Step 4, but the protocol is designed for scenarios without such benchmarks.

Output: Preprocessed dataset files (e.g., .csv or .xlsx) containing text instances, optional ground truth labels, and metadata. Store as {dataset_name}_prepared.xlsx.

Step 2: Model-based coding – code the same sample with LLMs. Initial internal reliability analysis (Krippendorffs' Alpha)

With the dataset prepared, deploy an ensemble of LLMs to generate independent labels for each instance, treating models as surrogate annotators. Ensemble diversity is crucial: selecting models from different developers, sizes depending on the task (small/medium to reduce costs), and training paradigms justifies the "crowd" analogy, as it captures varied perspectives and reduces correlated errors. We recommend 10–15 models; fewer risks insufficient evidence to benefit from crow aggregation, while more increases cost and computation without proportional gains.

Apply models in zero-shot mode (no examples in prompts) to simulate real-world scenarios without training data. Follow a standardized classification protocol like Domínguez-Díaz et al. (2025), which includes prompt templates from codebook, output validation (e.g., enforcing single-label responses), and retry logic for failures. Access via APIs (e.g., OpenAI, Google, Anthropic) with temperature = 0 for determinism. Prompts should incorporate the codebook from Step 1, specifying the task, labels, and output format (e.g., "Output only a single label: "). For challenging tasks, test prompt variants—e.g., detailed category descriptions vs. label names only—to evaluate robustness.

Post-coding, assess initial reliability with Krippendorff's alpha (nominal level), chosen for its flexibility with multiple annotators, missing data handling, and applicability to categorical labels. It provides a global measure of inter-LLM agreement, where α > 0.8 indicates strong reliability for proxy ground truth. Compute using the krippendorff Python package (version 0.7), with bootstrap resampling (n=1,000 samples) for confidence intervals, lower/upper bounds, standard errors, and variability assessment. This step justifies proceeding to aggregation only if baseline agreement is moderate (α > 0.6); otherwise, refine prompts or ensemble.

Output: Annotated dataset file with LLM-specific label columns; reliability metrics in krippendorff_alpha.csv (alpha, CI bounds, standard error, bootstrap n per dataset).

Step 3: Consensus building – aggregate outputs via majority vote.

Aggregation transforms individual labels into a collective consensus, approximating ground truth as a "likelihood surface" where high-vote labels are deemed more reliable. We selected simple majority vote for aggregation, as it counts the most frequent label per instance (ties resolved lexicographically). This method was chosen because: (1) it is conceptually simple and interpretable, aligning with classic "wisdom of crowds" principles by weighting all annotators equally unless skills vary dramatically; (2) it requires no training data or hyperparameters, suiting zero-shot ensembles; (3) prior annotation studies show it outperforms complex probabilistic models (e.g., Dawid-Skene, GLAD) in moderate ensembles (10–20 annotators) with relatively high individual quality, avoiding overfitting to noise (Paun et al., 2018; Tu et al., 2019; Majdi & Rodríguez, 2023); (4) it enables straightforward comparison with individual performances in validation.

Implement using crowd-kit (version 1.4.1; https://pypi.org/project/crowd-kit/), via MajorityVote on a DataFrame of LLM labels (rows: instances, columns: LLMs).

Output: Updated dataset file with majority_vote_label and majority_vote_consistency columns.

Step 4: Post-hoc analysis of the wisdom of the AI-CROWD – Post-Aggregation metrics

After obtaining the majority-vote consensus labels, the AI-CROWD protocol performs a diagnostic post-hoc evaluation of the AI crowd's behavior. This step is essential because the aggregated label is only an approximation to ground truth: its quality depends on the internal coherence and reliability of the ensemble. Without post-aggregation metrics, researchers would lack tools to assess whether the consensus is trustworthy, to identify systematic weaknesses in the AI crowd, or to decide which instances might require further scrutiny (e.g., human review). By quantifying both individual alignment and collective uncertainty, this step provides transparency and actionable insights, enabling users to interpret the approximation critically and refine the protocol iteratively. Two complementary metrics are computed: Annotator skill estimates and task uncertainty.

For each LLM, calculate its alignment with the majority-vote consensus as the simple accuracy (proportion of instances where the model's label matches the aggregated label). This annotator skill estimate metric serves as a proxy for how well each individual "annotator" (LLM in this case) represents the collective judgment of the crowd. High alignment indicates that a model is well-integrated into the ensemble and contributes reliably to the consensus; low alignment flags potential outliers whose predictions systematically diverge, possibly due to domain misalignment, prompt sensitivity, or idiosyncratic biases. Accuracy against consensus is computationally lightweight, interpretable, and directly comparable across models. It avoids circularity (since the consensus is derived from all models) while still providing a useful signal for ensemble pruning or weighting in future iterations. We compute point estimates and 95% bootstrap confidence intervals (1,000 resamples) using crowd-kit's MajorityVote aggregator estimated skills and custom bootstrapping code, yielding means, lower/upper bounds, and standard errors per model. Results are stored in llm_mvskills.csv.

Task uncertainty captures the level of collective disagreement, computed the average Shannon entropy of the per-instance label probability distributions induced by the ensemble of LLMs.

Entropy is calculated over the empirical distribution of votes (fraction of models assigning each label). To make the measure more epistemically meaningful, we weight each model's contribution by its estimated skill (alignment score from above): higher-skill models receive greater weight in the probability distribution. This transforms raw vote disagreement into a skill-weighted posterior uncertainty over the true label — disagreement among reliable models increases uncertainty far more than disagreement among unreliable ones. Shannon entropy is a standard information-theoretic measure of predictive dispersion; weighting by skill aligns with Bayesian principles and focuses attention on substantively important divergences rather than noise from weak annotators. Average task-level entropy (with 95% bootstrap CIs) reveals which datasets or tasks are inherently ambiguous for the AI crowd, guiding decisions about prompt refinement, model exclusion, or human-in-the-loop intervention. Results are stored in task_entropy_mvskills.csv (mean entropy, CI bounds, standard error).

These two metrics together form a diagnostic dashboard: high consensus alignment + low task uncertainty supports strong confidence in the aggregated labels; low alignment + high uncertainty signals the need for caution or further steps (bias mapping, human arbitration). This post-hoc layer is what distinguishes the protocol from naive majority voting — it does not blindly trust the crowd but actively interrogates its internal dynamics.

[Optional] Validation: Post-hoc aggregate–ground-truth accuracy

When a representative human-annotated ground truth set is available (e.g., the original test labels of benchmark datasets or a held-out golden subset of custom data), the protocol includes an optional external validation step. This step directly measures how well the majority-vote consensus — and, for comparison, each individual LLM — approximates the human gold standard, providing the strongest empirical evidence of the method's effectiveness in the specific validation context. We compute two standard classification metrics against ground truth:

- Accuracy: the proportion of instances where the predicted label (consensus or individual model) exactly matches the human label.

- Macro-averaged F1-score: the harmonic mean of precision and recall, averaged across classes without weighting by class size. Macro-F1 is preferred here because it treats all classes equally, which is important when datasets are imbalanced or when minority classes carry substantive meaning (e.g., rare citation intents).

Both metrics are calculated using scikit-learn functions (accuracy, F1-score), with 95% bootstrap confidence intervals (1,000 resamples) to account for sampling variability and provide uncertainty quantification. Results are stored in crowd_accuracy.csv and crowd_f1.csv.

External validation against human labels serves as the ultimate sanity check — it answers whether the AI crowd's approximation is practically useful in domains where ground truth exists. Including both individual models and the consensus allows direct comparison: if majority vote performs competitively with (or better than) the single best LLM while being more robust across tasks, this strongly supports the value of ensemble aggregation. Narrow bootstrap intervals in high-performing cases confirm estimate stability; wider intervals in difficult tasks (e.g., SciCite) highlight realistic limitations.

This optional step is not required for applying the protocol to datasets without ground truth (the typical real-world use case), but it is strongly recommended whenever such labels are accessible, as it provides the most convincing empirical justification for adopting the aggregated labels in downstream analyses.

**Method validation**

Step 1: Dataset preparation

To validate the proposed wisdom of the AI-CROWD protocol, we selected four well-established benchmark datasets (Table 1) that are widely used in machine learning and natural language processing research as standardized testbeds for text classification tasks. These datasets were deliberately chosen to cover a diverse range of classification problems, underlying knowledge types, and levels of task difficulty, thereby providing a robust evaluation of the protocol's generalizability. The selected datasets are:

- AG News (Zhang et al., 2015): A large-scale news topic classification task with four thematic categories (World, Sports, Business, Sci/Tech). It draws on latent world knowledge and journalistic discourse.

- IMDb (Maas et al., 2011): A binary sentiment analysis benchmark based on movie reviews, testing the models' ability to detect affective polarity from subjective, opinionated text.

- DBpedia-14 (Zhang et al., 2015): An ontological entity classification task with 14 non-overlapping classes derived from Wikipedia/DBpedia summaries (e.g., Company, Artist, Building, Album). It requires fine-grained factual and encyclopedic knowledge.

- SciCite (Cohan et al., 2019): A citation intent classification task in scientific literature, with three classes (background, method, result). This dataset evaluates interpretive reasoning over domain-specific academic discourse and is notably more challenging due to subtle contextual cues.

Detailed descriptions, including task types, number of classes, typical sizes, and original references, are provided in Table 1. These four datasets collectively represent substantial diversity in the nature of the classification target (thematic, sentiment, ontological, intent-based),the required type of knowledge (general/world knowledge, affective understanding, structured factual knowledge, scientific reasoning), the number of classes (ranging from 2 to 14), and the linguistic register and domain specificity (news, reviews, encyclopedic summaries, scientific prose).

Table 1. Description of Datasets Used in the Study

| Dataset | Task Type | Brief Description | # Classes | Class Labels | Size | Reference |
|---|---|---|---|---|---|---|
| AG News | Topic classification of news headlines | News headlines and descriptions classified into four thematic categories. | 4 | World, Sports, Business, Sci/Tech | 127,600 | Zhang, X., Zhao, J., & LeCun, Y. (2015) |
| DBpedia-14 | Ontological/ entity type classification | Wikipedia entity summaries classified into 14 non-overlapping ontological classes. | 14 | Company, EducationalInstitution, Artist, Athlete, OfficeHolder, MeanOfTransportation, Building, NaturalPlace, Village, Animal, Plant, Album, Film, WrittenWork | 630,000 | Zhang, X., Zhao, J., & LeCun, Y. (2015) |
| IMDb | Binary sentiment analysis | Movie reviews classified as expressing positive or negative sentiment. | 2 | Positive, Negative | 50,000 | Maas et al. (2011) |
| SciCite | Citation intent classification | Text spans from scientific papers containing a citation, classified by the intent/motivation of the citation. | 3 | Background, Method, Result | 11,020 | Cohan et al. (2019) |

Note. Dataset descriptions are based on standard benchmark usage as reported in the original sources. Dataset sizes refer to the standard benchmark splits used in the literature. For SciCite, the total number of annotated examples is reported, as the exact train/dev/test partition may vary slightly depending on the source.

Given the very large scale of most of these datasets, and considering the computational cost of labeling the full corpora with multiple LLMs, we adopted a sampling strategy for this methodological validation. From the standard test set of each dataset, we extracted a stratified random sample of 1,000 instances, preserving the original class distribution. This sample size balances computational feasibility with statistical power, ensuring that the resulting reliability and aggregation metrics remain representative of the full dataset behavior while allowing precise estimation of confidence intervals. The resulting 1,000-instance subsets were used as the input for all subsequent steps of the protocol (model-based coding, reliability analysis, aggregation, and post-hoc validation against ground truth). All sampling and preprocessing steps are fully reproducible and are included in the accompanying code repository.

Step 2: Model-based coding – initial reliability analysis

To generate the raw annotations for aggregation, the full representative samples from each dataset were independently coded by an ensemble of 11 LLMs in a zero-shot setting. The selected models were: deepseek-chat, gpt-5-nano, gpt-5-mini, gpt-5.1, gemini-2.5-flash-lite, gemini-2.5-flash, grok-4-1-fast-non-reasoning, claude-sonnet-4-5-20250929, claude-haiku-4-5-20251001, mistral-medium-2508, and mistral-small-2506. This ensemble was deliberately composed of the most recent available versions (as of December 2025) of small to medium-sized, non-reasoning (i.e., direct-inference) models released by major developers (OpenAI, Google, Anthropic, xAI, Mistral AI, and DeepSeek). The inclusion of models spanning different developers, parameter scales, training objectives, and architectural families aims to maximize diversity in the "AI-CROWD," thereby enhancing the robustness of the subsequent majority-vote aggregation and allowing the detection of systematic biases or convergence patterns that transcend individual model idiosyncrasies.

All models were applied following the automated LLM classification protocol previously described by Domínguez-Díaz et al. (2025), which specifies standardized zero-shot prompting, label mapping, output parsing, and error-handling procedures for closed-set multiclass tasks. The exact prompts used in this study are provided in Supplementary Material SM1. For the three benchmark datasets with clear categorical structures (AG News, IMDb, and DBpedia-14), a single detailed prompt was employed that included a brief task description, the full list of class labels, and explicit instructions to select exactly one label without additional commentary. For the SciCite dataset—given its greater interpretive complexity and the known sensitivity of citation intent classification to prompt formulation—we deliberately tested two distinct prompting strategies to assess their impact on ensemble agreement and downstream aggregation quality. The primary version included a comprehensive description of each citation intent category (background, method, result), including linguistic cues and contextual indicators typically associated with each class. A secondary abbreviated version (SciCite Prompt 2) provided only the names of the three categories without explanatory text, relying solely on the model's pre-existing semantic understanding of the terms "background," "method," and "result." The dual-prompt approach for SciCite serves both exploratory and methodological purposes: it allows evaluation of whether richer semantic scaffolding improves inter-model consensus (as reflected in Krippendorff's alpha and task entropy metrics) and helps quantify prompt-induced variability in the AI crowd's collective judgment on a challenging, domain-specific task.

All coding was performed using the official API endpoints available in December 2025, with reasoning parameters set to the lowest possible values (none or low reasoning) to maximize cost-efficiency and default values in the remaining parameters. Class labels were parsed from model responses and stored next to their corresponding instance IDs in a CSV file per dataset and model. Parsing failures and response refusals were classified as "response_error" and the resulting errors were logged for revision. Queries with erroneous responses were repeated when the logged error was unrelated to the instance content, such as reaching an API usage limit. When all the responses were recorded, an excel file was generated per dataset, containing the original sample data as well as one column per LLM with the parsed response. The resulting per-instance label predictions from the 11 LLMs constitute the input for the subsequent reliability and aggregation analyses (Steps 3 and 4).

Table 2 reports the overall inter-rater reliability (Krippendorff's Alpha) across all LLMs for each dataset, demonstrating high agreement on simpler classification tasks (AG News: α =

0.902; IMDb: α = 0.909; DBpedia-14: α = 0.928) but lower consistency on the more nuanced citation intent task (SciCite with Prompt 1: α = 0.681; SciCite with Prompt 2: α = 0.568). These patterns indicate that LLM ensembles can reliably approximate ground truth for thematic and sentiment classification but face challenges with intent-based labeling, potentially due to interpretive ambiguity in scientific text (see Table 1 for dataset details).

Table 2. Krippendorff's Alpha for Coding Across Datasets Using 11 LLMs

| Dataset | Krippendorff's Alpha | 95% CI Lower | 95% CI Upper | SE |
| --- | --- | --- | --- | --- |
| AG-News | 0.902 | 0.889 | 0.914 | 0.007 |
| IMDb | 0.909 | 0.893 | 0.924 | 0.008 |
| DBpedia-14 | 0.928 | 0.918 | 0.938 | 0.005 |
| SciCite | 0.681 | 0.660 | 0.702 | 0.011 |
| SciCite (Prompt 2) | 0.568 | 0.545 | 0.590 | 0.011 |

*Note*. All confidence intervals (CI) were calculated using bootstrap with 1000 samples and 95% confidence level.

Step 3: Consensus building – aggregate outputs via majority vote

Once the 11 LLMs had independently labeled each of the 1,000 sampled instances per dataset (Step 2), we proceeded to aggregate their individual predictions into a single consensus label for each instance. This aggregation step constitutes the core of the Wisdom of the AI-CROWD protocol, transforming a set of potentially divergent model outputs into a unified approximation to ground truth.

We employed simple majority vote as the primary aggregation method. For each instance and each dataset, the label that received the highest number of votes across the 11 LLMs was selected as the aggregated (consensus) label. In the case of ties (two or more labels receiving the same maximum number of votes), the tie was resolved by selecting the label with the lowest lexicographical order among the tied options—a deterministic and reproducible rule that has negligible impact given the low frequency of exact ties in practice (observed in 0.44% of instances across all datasets; 0% for AG-News, IMDb and DBpedia-14, 0.5% for SciCite, and 1.44% for SciCite with Prompt2). The majority-vote aggregation was implemented using the crowd-kit Python library (version 1.2.1; https://pypi.org/project/crowd-kit/), specifically the MajorityVote class from the crowdkit.aggregation.classification.majority_vote module. This implementation ensures efficient computation, handles multi-class problems natively, and provides optional consistency scores for each aggregated prediction (the normalized margin between the winning vote count and the second-highest count).

This step produced, for each of the five labeling conditions (AG News, IMDb, DBpedia-14, SciCite, SciCite Prompt 2), a fully annotated dataset in which every instance now carries both the individual LLM predictions and a single aggregated label derived from majority voting. All aggregation runs were executed in a reproducible Python environment, with fixed random seeds where applicable.

Step 4: Post-hoc analysis of the wisdom of the AI-CROWD– Post-Aggregation metrics

To understand how individual LLMs contribute to this collective judgment, Tables 3–7 present the alignment of each LLM with the majority-vote ensemble consensus (i.e., annotator skill estimates), along with 95% bootstrap confidence intervals. High alignment indicates strong congruence with the crowd consensus, while lower values highlight potential outliers or models less representative of the group judgment.

On AG News (news topic classification; Table 3), alignment ranged from 0.897 (gpt-5-nano) to 0.974 (grok-4-1-fast-non-reasoning), with a tight cluster of frontier models (grok-4-1-fast-non-reasoning, claude-haiku-4-5-20251001, gpt-5.1) exceeding 0.96 agreement with the ensemble. This reflects robust collective understanding of straightforward news domains. For IMDb (binary sentiment classification; Table 4), alignment was exceptionally high overall, ranging from 0.935 (gemini-2.5-flash-lite) to 0.991 (mistral-medium-2508). Several models (mistral-medium-2508, claude-sonnet-4-5-20250929, gpt-5.1, grok-4-1-fast-non-reasoning) showed near-perfect congruence (>0.98) with the majority vote, suggesting strong crowd convergence on polarity detection. DBpedia-14 (entity type classification into 14 ontological classes; Table 5) exhibited the highest alignments, with top models reaching 0.994–0.996 (claude-sonnet-4-5-20250929, mistral-medium-2508, gpt-5.1). The narrow gaps among leading models underscore the ensemble's stability on knowledge-rich entity-typing tasks.

In contrast, the SciCite dataset (citation intent classification; Table 6) showed greater dispersion in alignment (0.760 for mistral-medium-2508 to 0.925 for gemini-2.5-flash), highlighting interpretive challenges in scientific contexts. Gemini-2.5-flash, gpt-5.1, and claude-haiku-4-5-20251001 aligned most closely with the consensus. The alternative prompt version (SciCite Prompt 2; Table 7) produced slightly lower but similar patterns (range: 0.696–0.889), with gemini-2.5-flash again showing the strongest alignment to the ensemble.

Across all datasets, models such as gemini-2.5-flash, gpt-5.1, grok-4-1-fast-non-reasoning, and certain Claude/Mistral variants consistently ranked highest in alignment with the majority-vote consensus. Bootstrap confidence intervals were narrow for high-alignment cases, confirming stable estimates. These results demonstrate that while the ensemble achieves high overall agreement (Table 2), individual alignment metrics reveal the degree of internal coherence within the AI crowd, enabling identification of outlier LLMs and supporting the protocol's value for scalable, reliable ground truth approximation in content analysis — particularly when combined with prompt refinement for ambiguous tasks.

Table 3. Alignment of Individual LLMs with Majority-Vote Ensemble on the AG News Dataset

| Rank | LLM | Mean Skill | 95% CI |
|---|---|---|---|
| 1 | grok-4-1-fast-non-reasoning | 0.974 | [0.967, 0.982] |
| 2 | claude-haiku-4-5-20251001 | 0.969 | [0.961, 0.978] |
| 3 | gpt-5.1 | 0.968 | [0.961, 0.977] |
| 4 | mistral-small-2506 | 0.966 | [0.957, 0.975] |
| 5 | gemini-2.5-flash-lite | 0.959 | [0.949, 0.969] |
| 6 | gpt-5-mini | 0.957 | [0.948, 0.967] |
| 7 | mistral-medium-2508 | 0.953 | [0.943, 0.963] |
| 8 | claude-sonnet-4-5-20250929 | 0.952 | [0.942, 0.962] |
| 9 | deepseek-chat | 0.938 | [0.927, 0.949] |
| 10 | gemini-2.5-flash | 0.936 | [0.924, 0.948] |
| 11 | gpt-5-nano | 0.897 | [0.882, 0.911] |

*Note*. All confidence intervals (CI) were calculated using 1000 bootstrap samples at the 95% confidence level.

Table 4. Alignment of Individual LLMs with Majority-Vote Ensemble on the IMDb Dataset

| Rank | LLM | Mean Skill | 95% CI |
|---|---|---|---|
| 1 | mistral-medium-2508 | 0.991 | [0.987, 0.995] |
| 2 | claude-sonnet-4-5-20250929 | 0.989 | [0.984, 0.994] |
| 3 | gpt-5.1 | 0.988 | [0.983, 0.994] |
| 4 | grok-4-1-fast-non-reasoning | 0.988 | [0.983, 0.994] |
| 5 | gpt-5-mini | 0.986 | [0.981, 0.991] |
| 6 | claude-haiku-4-5-20251001 | 0.985 | [0.979, 0.991] |
| 7 | gpt-5-nano | 0.98 | [0.973, 0.987] |
| 8 | mistral-small-2506 | 0.958 | [0.949, 0.968] |
| 9 | deepseek-chat | 0.943 | [0.932, 0.954] |
| 10 | gemini-2.5-flash | 0.939 | [0.929, 0.951] |
| 11 | gemini-2.5-flash-lite | 0.935 | [0.923, 0.947] |

*Note*. All confidence intervals (CI) were calculated using 1000 bootstrap samples at the 95% confidence level.

Table 5. Alignment of Individual LLMs with Majority-Vote Ensemble on the DBpedia-14 Dataset

| Rank | LLM | Mean Skill | 95% CI |
|---|---|---|---|
| 1 | claude-sonnet-4-5-20250929 | 0.996 | [0.994, 0.998] |
| 2 | mistral-medium-2508 | 0.994 | [0.991, 0.998] |
| 3 | gpt-5.1 | 0.994 | [0.99, 0.997] |
| 4 | claude-haiku-4-5-20251001 | 0.989 | [0.984, 0.994] |
| 5 | gpt-5-mini | 0.986 | [0.981, 0.992] |
| 6 | mistral-small-2506 | 0.982 | [0.976, 0.989] |
| 7 | grok-4-1-fast-non-reasoning | 0.978 | [0.971, 0.986] |
| 8 | deepseek-chat | 0.899 | [0.886, 0.914] |
| 9 | gpt-5-nano | 0.899 | [0.885, 0.913] |
| 10 | gemini-2.5-flash | 0.897 | [0.884, 0.912] |
| 11 | gemini-2.5-flash-lite | 0.895 | [0.881, 0.91] |

*Note*. All confidence intervals (CI) were calculated using 1000 bootstrap samples at the 95% confidence level.

Table 6. Alignment of Individual LLMs with Majority-Vote Ensemble on the SciCite (Prompt1) Dataset

| Rank | LLM | Mean Skill | 95% CI |
|---|---|---|---|
| 1 | gemini-2.5-flash | 0.925 | [0.913, 0.938] |
| 2 | gpt-5.1 | 0.905 | [0.893, 0.92] |
| 3 | claude-haiku-4-5-20251001 | 0.903 | [0.888, 0.917] |
| 4 | grok-4-1-fast-non-reasoning | 0.897 | [0.883, 0.911] |
| 5 | gemini-2.5-flash-lite | 0.885 | [0.869, 0.901] |
| 6 | deepseek-chat | 0.88 | [0.865, 0.895] |
| 7 | claude-sonnet-4-5-20250929 | 0.855 | [0.838, 0.872] |
| 8 | gpt-5-mini | 0.845 | [0.828, 0.861] |
| 9 | mistral-small-2506 | 0.83 | [0.812, 0.847] |
| 10 | gpt-5-nano | 0.802 | [0.784, 0.821] |
| 11 | mistral-medium-2508 | 0.760 | [0.739, 0.781] |

Table 7. Alignment of Individual LLMs with Majority-Vote Ensemble on the SciCite (Prompt 2) Dataset

| Rank | LLM | Mean Skill | 95% CI |
|---|---|---|---|
| 1 | gemini-2.5-flash | 0.889 | [0.875, 0.904] |
| 2 | grok-4-1-fast-non-reasoning | 0.878 | [0.863, 0.894] |
| 3 | deepseek-chat | 0.878 | [0.863, 0.893] |
| 4 | gpt-5-mini | 0.863 | [0.847, 0.879] |
| 5 | gpt-5.1 | 0.854 | [0.837, 0.87] |
| 6 | claude-haiku-4-5-20251001 | 0.846 | [0.829, 0.863] |
| 7 | claude-sonnet-4-5-20250929 | 0.792 | [0.773, 0.811] |
| 8 | gemini-2.5-flash-lite | 0.771 | [0.752, 0.79] |
| 9 | mistral-small-2506 | 0.753 | [0.733, 0.774] |
| 10 | gpt-5-nano | 0.7 | [0.678, 0.722] |
| 11 | mistral-medium-2508 | 0.696 | [0.674, 0.717] |

*Note*. All confidence intervals (CI) were calculated using 1000 bootstrap samples at the 95% confidence level.

Task uncertainty was assessed by computing the Shannon entropy of the label probability distribution induced by the 11 LLMs for each instance in the datasets, with the mean entropy per dataset serving as an indicator of overall collective disagreement or ambiguity (Table 8). Lower mean entropy indicates high consensus among the LLMs (i.e., most models assign very similar label probabilities), whereas higher values signal substantial divergence in their predictions. The results reveal clear differences across tasks. On the relatively straightforward classification problems—AG News (mean entropy = 0.388), IMDb (0.205), and DBpedia-14 (0.442)—the ensemble exhibits low to moderate uncertainty. The particularly low entropy on IMDb aligns with the binary nature of sentiment classification and the strong convergence observed in both overall inter-LLM agreement (Krippendorff's α = 0.909 in Table 2) and individual alignment with the majority-vote consensus (Table 4). These findings suggest that, for tasks with clear categorical boundaries and large training signals in pre-training corpora, the 11-LLM crowd reaches a high degree of internal coherence. In contrast, the citation intent classification tasks (SciCite and SciCite with Prompt 2) display markedly higher uncertainty, with mean entropies of 1.108 and 1.447, respectively. These elevated values reflect substantial disagreement among LLMs when interpreting the motivation behind a citation in scientific text, consistent with the lower overall Krippendorff's alpha values (0.681 and 0.568) and the wider spread in individual alignment scores (Tables 6 and 7). The even higher uncertainty with the alternative prompt (SciCite Prompt 2) further indicates that prompt formulation can significantly influence ensemble divergence on nuanced, context-dependent tasks.

Table 8. Task Uncertainty (Mean Shannon Entropy of LLM Label Distributions) Across Datasets

| Dataset | Mean Entropy | 95% CI | Std. Error |
| --- | --- | --- | --- |
| AG News | 0.388 | [0.376, 0.400] | 0.006 |
| IMDb | 0.205 | [0.196, 0.214] | 0.005 |
| DBpedia-14 | 0.442 | [0.429, 0.456] | 0.007 |
| SciCite | 1.108 | [1.092, 1.123] | 0.008 |
| SciCite (Prompt 2) | 1.447 | [1.432, 1.461] | 0.008 |

*Note*. All confidence intervals (CI) were calculated using 1000 bootstrap samples at the 95% confidence level. Higher entropy reflects greater disagreement among LLMs regarding the predicted label distribution for each instance.

Overall, the task uncertainty results complement the inter-rater agreement (Table 2) and individual alignment analyses (Tables 3–7) by providing a direct measure of predictive dispersion within the AI crowd. Low-entropy tasks correspond to domains where majority-vote aggregation is likely to produce highly reliable proxy ground truth labels, while high-entropy tasks highlight the need for additional strategies (e.g., prompt refinement, model filtering, or hybrid human-AI approaches) to mitigate ambiguity in large-scale content analysis. Bootstrap confidence intervals are narrow across all datasets, confirming the stability of these uncertainty estimates.

The two post-aggregation metrics—annotator alignment with consensus and skill-weighted task uncertainty—together function as a diagnostic dashboard for evaluating the reliability of the AI crowd's approximation. High consensus alignment combined with low task uncertainty (as observed on structured tasks like DBpedia-14 and IMDb) provides strong confidence that the majority-vote labels constitute a robust approximation for ground truth. Conversely, lower alignment or elevated uncertainty (particularly evident on SciCite tasks) signals caution and highlights the need for further steps such as bias mapping, targeted human arbitration on high-entropy instances, or prompt/model refinement. This protocol validation shows that the diagnostic layer fundamentally distinguishes the present protocol from naive majority voting: rather than blindly accepting the crowd's verdict, the framework actively interrogates its internal dynamics, offering researchers transparent criteria to judge when the aggregated approximation can be used directly and when it should be treated with skepticism or supplemented. In doing so, it advances a more reflexive and accountable approach to LLM-based content analysis at scale.

Post-hoc aggregate–ground-truth accuracy

To validate the protocol's effectiveness in approximating ground truth, we evaluated the agreement between the LLM outputs (individual models and majority-vote aggregation) and the available human-annotated ground truth labels on representative golden sets for each dataset. Tables 9–13 report mean accuracy and macro-averaged F1-scores (with 95% bootstrap confidence intervals) against these ground truth benchmarks, providing a direct measure of labeling quality.

Across datasets, individual LLMs exhibited strong but variable performance, with frontier models like gemini-2.5-flash, claude-sonnet-4-5-20250929, and gpt-5.1 frequently ranking highest. For instance, on AG News (Table 9), gemini-2.5-flash achieved the top F1-score (0.913) and accuracy (0.913), outperforming others in this four-class topic classification task. Similarly, on DBpedia-14 (Table 10), gpt-5.1 led with near-perfect scores (F1 = 0.987, accuracy = 0.987), reflecting robust entity typing capabilities. For IMDb's binary sentiment analysis (Table 11), claude-sonnet-4-5-20250929 excelled (F1 = 0.961, accuracy = 0.961). On the more challenging citation intent tasks, gemini-2.5-flash again dominated SciCite (F1 = 0.819, accuracy = 0.816; Table 12) and SciCite (Prompt 2) (F1 = 0.767, accuracy = 0.763; Table 13), highlighting its strength in nuanced scientific interpretation.

Importantly, the majority-vote aggregation consistently performed competitively, often matching or approaching the best individual LLMs while surpassing lower performers. On structured tasks like DBpedia-14 (F1 = 0.985 vs. top 0.987) and IMDb (F1 = 0.952 vs. top 0.961), majority vote ranked in the top three, demonstrating the "wisdom of the AI crowd" in leveraging ensemble diversity to achieve high reliability. Even on AG News (F1 = 0.874 vs. top 0.913), it ranked fourth, outperforming seven individual models and providing a balanced approximation. For SciCite (F1 = 0.791 vs. top 0.819) and SciCite (Prompt 2) (F1 = 0.757 vs. top 0.767), majority vote placed fourth and second, respectively, closely trailing the leaders despite greater task ambiguity (as evidenced by higher uncertainty in Table 8). This pattern suggests that while majority vote may not always exceed the single best LLM, it robustly approximates ground truth by mitigating individual biases and errors—particularly valuable for large-scale datasets where selecting the "best" model a priori is challenging. Bootstrap intervals were narrow for high performers, affirming estimate stability, and these validation results support the protocol's pragmatic utility in computational social science.

Table 9. Performance Against Ground Truth (Accuracy and F1-Score) for Individual LLMs and Majority Vote on the AG News Dataset

| Rank | Model | Mean Accuracy | 95% CI (Acc) | Mean F1 | 95% CI (F1) |
|---|---|---|---|---|---|
| 1 | gemini-2.5-flash | 0.913 | [0.895, 0.93] | 0.913 | [0.895, 0.93] |
| 2 | claude-sonnet-4-5-20250929 | 0.904 | [0.884, 0.923] | 0.904 | [0.885, 0.921] |
| 3 | mistral-medium-2508 | 0.886 | [0.866, 0.906] | 0.887 | [0.868, 0.905] |
| 4 | MajorityVote | 0.874 | [0.852, 0.894] | 0.874 | [0.852, 0.893] |
| 5 | mistral-small-2506 | 0.872 | [0.851, 0.892] | 0.872 | [0.851, 0.892] |
| 6 | gpt-5.1 | 0.864 | [0.842, 0.884] | 0.861 | [0.839, 0.882] |
| 7 | grok-4-1-fast-non-reasoning | 0.862 | [0.842, 0.883] | 0.861 | [0.837, 0.883] |
| 8 | gpt-5-mini | 0.855 | [0.832, 0.877] | 0.853 | [0.832, 0.874] |
| 9 | gemini-2.5-flash-lite | 0.854 | [0.832, 0.876] | 0.852 | [0.829, 0.874] |
| 10 | claude-haiku-4-5-20251001 | 0.846 | [0.823, 0.867] | 0.844 | [0.82, 0.866] |
| 11 | deepseek-chat | 0.828 | [0.803, 0.853] | 0.822 | [0.795, 0.846] |
| 12 | gpt-5-nano | 0.814 | [0.79, | 0.815 | [0.79, 0.837] |

| | | | 0.838] | | |

Table 10. Performance Against Ground Truth (Accuracy and F1-Score) for Individual LLMs and Majority Vote on the DBpedia-14 Dataset

| Rank | Model | Mean Accuracy | 95% CI (Acc) | Mean F1 | 95% CI (F1) |
|---|---|---|---|---|---|
| 1 | gpt-5.1 | 0.987 | [0.979, 0.993] | 0.987 | [0.98, 0.993] |
| 2 | claude-sonnet-4-5-20250929 | 0.985 | [0.977, 0.992] | 0.985 | [0.977, 0.992] |
| 3 | MajorityVote | 0.985 | [0.978, 0.992] | 0.985 | [0.976, 0.992] |
| 4 | mistral-medium-2508 | 0.979 | [0.97, 0.987] | 0.979 | [0.97, 0.987] |
| 5 | gpt-5-mini | 0.975 | [0.966, 0.984] | 0.978 | [0.969, 0.987] |
| 6 | claude-haiku-4-5-20251001 | 0.977 | [0.967, 0.986] | 0.977 | [0.968, 0.985] |
| 7 | grok-4-1-fast-non-reasoning | 0.971 | [0.961, 0.98] | 0.971 | [0.96, 0.981] |
| 8 | mistral-small-2506 | 0.97 | [0.959, 0.98] | 0.971 | [0.96, 0.981] |
| 9 | gpt-5-nano | 0.897 | [0.878, 0.916] | 0.907 | [0.888, 0.925] |
| 10 | gemini-2.5-flash | 0.887 | [0.867, 0.906] | 0.888 | [0.867, 0.906] |
| 11 | deepseek-chat | 0.887 | [0.867, 0.907] | 0.887 | [0.869, 0.905] |
| 12 | gemini-2.5-flash-lite | 0.885 | [0.866, 0.904] | 0.886 | [0.867, 0.904] |

*Note*. All confidence intervals (CI) were calculated using 1000 bootstrap samples at the 95% confidence level.

Table 11. Performance Against Ground Truth (Accuracy and F1-Score) for Individual LLMs and Majority Vote on the IMDb Dataset

| Rank | Model | Mean Accuracy | 95% CI (Acc) | Mean F1 | 95% CI (F1) |
|---|---|---|---|---|---|
| 1 | claude-sonnet-4-5-20250929 | 0.961 | [0.949, 0.973] | 0.961 | [0.947, 0.972] |
| 2 | gpt-5.1 | 0.954 | [0.942, 0.967] | 0.954 | [0.941, 0.966] |
| 3 | MajorityVote | 0.952 | [0.939, 0.965] | 0.952 | [0.937, 0.964] |
| 4 | gpt-5-mini | 0.95 | [0.936, 0.963] | 0.95 | [0.936, 0.962] |
| 5 | mistral-medium-2508 | 0.949 | [0.936, 0.962] | 0.949 | [0.935, 0.962] |
| 6 | gpt-5-nano | 0.948 | [0.934, | 0.948 | [0.934, |

| | | | 0.961] | | 0.961] |
|---|---|---|---|---|---|
| 7 | grok-4-1-fast-non-reasoning | 0.944 | [0.93, 0.958] | 0.945 | [0.929, 0.959] |
| 8 | mistral-small-2506 | 0.94 | [0.925, 0.953] | 0.94 | [0.924, 0.954] |
| 9 | claude-haiku-4-5-20251001 | 0.939 | [0.924, 0.953] | 0.939 | [0.924, 0.953] |
| 10 | deepseek-chat | 0.903 | [0.884, 0.921] | 0.903 | [0.885, 0.921] |
| 11 | gemini-2.5-flash | 0.903 | [0.885, 0.92] | 0.903 | [0.884, 0.921] |
| 12 | gemini-2.5-flash-lite | 0.901 | [0.883, 0.918] | 0.9 | [0.883, 0.919] |

*Note*. All confidence intervals (CI) were calculated using 1000 bootstrap samples at the 95% confidence level.

Table 12. Performance Against Ground Truth (Accuracy and F1-Score) for Individual LLMs and Majority Vote on the SciCite Dataset

| Rank | Model | Mean Accuracy | 95% CI (Acc) | Mean F1 | 95% CI (F1) |
|---|---|---|---|---|---|
| 1 | gemini-2.5-flash | 0.816 | [0.793, 0.839] | 0.819 | [0.795, 0.843] |
| 2 | gpt-5.1 | 0.807 | [0.782, 0.83] | 0.811 | [0.787, 0.835] |
| 3 | claude-haiku-4-5-20251001 | 0.792 | [0.767, 0.816] | 0.795 | [0.771, 0.819] |
| 4 | MajorityVote | 0.784 | [0.759, 0.809] | 0.791 | [0.765, 0.816] |
| 5 | gemini-2.5-flash-lite | 0.778 | [0.754, 0.802] | 0.779 | [0.753, 0.804] |
| 6 | grok-4-1-fast-non-reasoning | 0.734 | [0.707, 0.761] | 0.747 | [0.721, 0.773] |
| 7 | gpt-5-nano | 0.749 | [0.724, 0.776] | 0.747 | [0.719, 0.772] |
| 8 | gpt-5-mini | 0.738 | [0.708, 0.766] | 0.74 | [0.711, 0.767] |
| 9 | claude-sonnet-4-5-20250929 | 0.715 | [0.688, 0.742] | 0.729 | [0.706, 0.755] |
| 10 | deepseek-chat | 0.713 | [0.684, 0.743] | 0.726 | [0.7, 0.752] |
| 11 | mistral-small-2506 | 0.668 | [0.638, 0.697] | 0.685 | [0.655, 0.712] |
| 12 | mistral-medium-2508 | 0.599 | [0.569, 0.632] | 0.625 | [0.595, 0.654] |

*Note*. All confidence intervals (CI) were calculated using 1000 bootstrap samples at the 95% confidence level.

Table 13. Performance Against Ground Truth (Accuracy and F1-Score) for Individual LLMs and Majority Vote on the SciCite (Prompt 2) Dataset

| Rank | Model | Mean Accuracy | 95% CI (Acc) | Mean F1 | 95% CI (F1) |
|---|---|---|---|---|---|
| 1 | gemini-2.5-flash | 0.763 | [0.737, 0.79] | 0.767 | [0.741, 0.792] |
| 2 | MajorityVote | 0.749 | [0.724, 0.775] | 0.757 | [0.732, 0.782] |
| 3 | grok-4-1-fast-non-reasoning | 0.742 | [0.714, 0.769] | 0.751 | [0.725, 0.778] |
| 4 | deepseek-chat | 0.723 | [0.694, 0.75] | 0.734 | [0.707, 0.759] |
| 5 | gpt-5.1 | 0.706 | [0.678, 0.732] | 0.727 | [0.701, 0.754] |
| 6 | claude-haiku-4-5-20251001 | 0.723 | [0.695, 0.75] | 0.723 | [0.692, 0.751] |
| 7 | gpt-5-mini | 0.694 | [0.664, 0.722] | 0.702 | [0.675, 0.728] |
| 8 | gemini-2.5-flash-lite | 0.693 | [0.668, 0.721] | 0.689 | [0.661, 0.719] |
| 9 | claude-sonnet-4-5-20250929 | 0.669 | [0.64, 0.699] | 0.682 | [0.654, 0.707] |
| 10 | gpt-5-nano | 0.651 | [0.621, 0.681] | 0.625 | [0.59, 0.655] |
| 11 | mistral-small-2506 | 0.581 | [0.551, 0.61] | 0.595 | [0.564, 0.626] |
| 12 | mistral-medium-2508 | 0.538 | [0.507, 0.57] | 0.552 | [0.518, 0.585] |

*Note*. All confidence intervals (CI) were calculated using 1000 bootstrap samples at the 95% confidence level.

**Limitations**

This paper introduces the AI-CROWD protocol, a replicable, zero-shot ensemble method based on large language models (LLMs) for approximating ground truth in large-scale content analysis when extensive human annotation is infeasible. The protocol operationalizes the aggregation of independent model outputs through majority-vote consensus across 11 diverse LLMs, explicitly treating the resulting labels as an approximation rather than as equivalents of ground truth. In doing so, the study offers a practical alternative to both large-scale human coding and single-model content analyses in contexts characterized by data volume, resource constraints, or the absence of reliable gold standards.

Several limitations of the proposed AI-CROWD protocol should be acknowledged. First, the methodology depends on the availability, stability, and policies of commercial or third-party LLM APIs. Although new actors and models are likely to emerge, this dependence on external infrastructures remains an inherent constraint of contemporary LLM-based research. Second, cost and computational scalability represent a practical limitation. Labeling even a relatively modest sample of 1,000 instances with an ensemble of 11 LLMs already entails non-trivial API expenses, particularly for higher-cost models. Third, the protocol is sensitive to prompt formulation, and systematic prompt optimization was not a primary focus of the present study. Although one or two prompt variants were tested for certain tasks, more extensive prompt engineering could plausibly improve judgements.

Fourth, the aggregation strategy assumes equal weighting of all models through simple majority voting. While this choice favors interpretability, it may be suboptimal in tasks characterized by high skill variance across models, as observed in the citation intent classification task. Skill-weighted aggregation procedures may yield improved results in such contexts and represent a natural extension of the protocol. Fifth, empirical validation is restricted to English-language, well-established benchmark datasets with relatively clean and high-quality text. The performance and diagnostic behavior of the protocol on noisier real-world data remain untested and warrant further investigation. Sixth, the use of stratified samples of 1,000 instances for validation, while justified by cost and feasibility considerations, limits the precision of bootstrap confidence intervals, particularly for rare classes. Eighth, the study does not assess temporal robustness. All results are based on model versions available at a specific point in time, and LLM behavior may drift due to retraining, fine-tuning, or safety updates. The stability of the AI-CROWD protocol across model versions or over longer temporal horizons therefore remains an open question.

Finally, it is important to recognize that human-annotated ground truth itself is imperfect. The benchmark labels used for external validation may contain errors. Consequently, the protocol's approximation is evaluated against an imperfect reference rather than an absolute standard of truth, reinforcing the paper's central claim that both human and AI-based annotations should be treated as probabilistic and diagnostically interrogated rather than taken at face value.

**Conclusion**

The AI-CROWD protocol represents a pragmatic advancement in large-scale content analysis, harnessing the "wisdom of the AI crowd" through ensemble aggregation of diverse LLMs to approximate ground truth where human benchmarks are infeasible. By generating a consensus-based likelihood surface—grounded in majority voting and interrogated via diagnostic metrics like Krippendorff's alpha, annotator alignment, and skill-weighted task uncertainty—the protocol offers researchers a transparent proxy for analytical confidence and validity.

Validation across four benchmarks demonstrates its efficacy: majority-vote approximations achieve competitive macro-F1 scores (0.757–0.985) against human labels, often rivaling or surpassing individual frontier models while mitigating biases through collective diversity. On structured tasks (e.g., DBpedia-14), high agreement and low uncertainty affirm robust reliability; on interpretive ones (e.g., SciCite), diagnostics flag prompt sensitivity and ambiguity, guiding refinements like human arbitration.

The implications are profound: AI-CROWD democratizes scalable labeling, reducing costs and enabling inquiry into massive datasets without sacrificing methodological rigor. Unlike naive voting, it fosters reflexive practice—actively probing crowd dynamics to distinguish signal from noise. Future extensions could incorporate bias mapping or hybrid human-AI workflows, further enhancing its utility in computational social science and beyond. Ultimately, by crowdsourcing wisdom from AI, the protocol empowers researchers to navigate data abundance with greater accountability and insight.

**CRediT author statement**

Luis de-Marcos: Conceptualization, Methodology, Writing- Original draft preparation, Supervision, Resources

Manuel Goyanes: Formal analysis, Methodology, Data curation, Validation, Writing- Reviewing and Editing

Adrián Domínguez-Díaz: Methodology, Software, Visualization, Investigation, Writing- Reviewing and Editing

# References

Araujo, T., Lock, I., & van de Velde, B. (2020). Automated visual content analysis (AVCA) in communication research: A protocol for large-scale image classification with pre-trained computer vision models. *Communication Methods and Measures, 14*(4), 239–265.

Baden, C., Kligler-Vilenchik, N., & Yarchi, M. (2020). Hybrid content analysis: Toward a strategy for the theory-driven, computer-assisted classification of large text corpora. *Communication Methods and Measures, 14*(3), 165–183.

Budak, C., Garrett, R. K., & Sude, D. (2021). Better crowdcoding: Strategies for promoting accuracy in crowdsourced content analysis. *Communication Methods and Measures, 15*(2), 141–155.

Cohan, A., Ammar, W., van Zuylen, M., & Cady, F. (2019). Structural scaffolds for citation intent classification in scientific publications. In *Proceedings of the 2019 Conference of the North American Chapter of the Association for Computational Linguistics: Human Language Technologies* (pp. 3586–3596).

Domínguez-Díaz, A., Goyanes, M., & de-Marcos, L. (2025). Automating content analysis of scientific abstracts using ChatGPT: A methodological protocol and use case. *MethodsX, 15*, 103431. https://doi.org/10.1016/j.mex.2025.103431

Dumitrache, A. (2015). Crowdsourcing Disagreement for Collecting Semantic Annotation. *European Semantic Web Conference* (pp. 701–710). Springer, Cham. https://doi.org/10.1007/978-3-319-18818-8_43

Dumitrache, A., Inel, O., Timmermans, B., Ortiz, C., Sips, R.-J., Aroyo, L., & Welty, C. (2018). Empirical Methodology for Crowdsourcing Ground Truth. *Semantic Web: – Interoperability, Usability, Applicability*. 2020;12(3):403-421. https://doi.org/10.3233/SW-200415

Goyanes, M., De-Marcos, L., & Domínguez-Díaz, A. (2024). Automatic gender detection: A methodological procedure and recommendations to computationally infer the gender from names with ChatGPT and gender APIs. *Scientometrics, 129*(11), 6867–6888.

Goyanes, M., & Piñeiro Naval, V. (2024). Análisis de contenido en SPSS y KALPHA: Procedimiento para un análisis cuantitativo fiable con la Kappa de Cohen y el Alpha de Krippendorff. *Estudios sobre el Mensaje Periodístico, 30*(1), 123–140.

Joo, J., & Steinert-Threlkeld, Z. C. (2022). Image as data: Automated content analysis for visual presentations of political actors and events. *Computational Communication Research, 4*(1).

Krippendorff, K. (2008). Systematic and random disagreement and the reliability of nominal data. *Communication Methods and Measures, 2*(4), 323–338.

Krippendorff, K. (2018). Content analysis: An introduction to its methodology. Sage Publications.

Li, J., Baba, Y., & Kashima, H. (2018). Incorporating Worker Similarity for Label Aggregation in Crowdsourcing. *Artificial Neural Networks and Machine Learning – ICANN 2018* (pp. 596–606). Springer, Cham. https://doi.org/10.1007/978-3-030-01421-6_57

Lind, F., Gruber, M., & Boomgaarden, H. G. (2017). Content analysis by the crowd: Assessing the usability of crowdsourcing for coding latent constructs. *Communication Methods and Measures, 11*(3), 191–209.

Maas, A. L., Daly, R. E., Pham, P. T., Huang, D., Ng, A. Y., & Potts, C. (2011). Learning word vectors for sentiment analysis. In *Proceedings of the 49th Annual Meeting of the Association for Computational Linguistics: Human Language Technologies* (pp. 142–150).

Majdi, M. S., & Rodríguez, J. J. (2023). Crowd-Certain: Label aggregation in crowdsourced and ensemble learning classification. *Research Square*. https://doi.org/10.21203/rs.3.rs-3488150/v1

Nishio, T., Ito, H., Tamura, T., & Morishima, A. (2025). Rationale-aware Label Aggregation in Crowdsourcing. *Proceedings of the ACM Collective Intelligence Conference* (pp. 188–197). https://doi.org/10.1145/3715928.3737480

Paun, S., Carpenter, B., Chamberlain, J., Hovy, D., Kruschwitz, U., & Poesio, M. (2018). Comparing Bayesian models of annotation. *Transactions of the Association for Computational Linguistics, 6*, 571–585. https://doi.org/10.1162/TACL_A_00040

Tu, J., Yu, G., Wang, J., Domeniconi, C., & Zhang, X. (2019). Attention-aware answers of the crowd. In *Proceedings of the SIAM International Conference on Data Mining* (pp. 451–459).

Zhang, J., Wu, X., & Sheng, V. S. (2016). Learning from crowdsourced labeled data: a survey. *Artificial Intelligence Review*, 46(4), 543–576. https://doi.org/10.1007/S10462-016-9491-9

Zhang, X., Zhao, J., & LeCun, Y. (2015). Character-level convolutional networks for text classification. In *Advances in Neural Information Processing Systems 28 (NeurIPS 2015)*.

# Supplementary Material 1 (SM1). Prompts for the different datasets.

**Prompt for the "AG-News" dataset**

```
You will have the role of a researcher that performs content
analysis tasks.

I need you to code a series of variables based on news
articles. Next, I will describe you the variables.

Variable "category" which represents the primary topic or
theme of the news article. Answer with one of the following
values: World, Sports, Business, Sci/Tech

All the variables are defined. This is the content to analyze
(between triple quotes):

"""{text}"""

Return the response in JSON format with a dictionary with a
key/value pair for each variable. Do not include any other
content in the response.
```

**Prompt for the "IMDb" dataset**

```
You will have the role of a researcher that performs content
analysis tasks.

I need you to code a series of variables based on encyclopedia
entries. Next, I will describe you the variables.

Variable "ontology_class" which represents the primary
ontological category or type of the entity described in the
text. Answer with one of the following values: Company,
EducationalInstitution, Artist, Athlete, OfficeHolder,
MeanOfTransportation, Building, NaturalPlace, Village, Animal,
Plant, Album, Film, WrittenWork

All the variables are defined. This is the content to analyze
(between triple quotes):

"""{text}"""

Return the response in JSON format with a dictionary with a
key/value pair for each variable. Do not include any other
content in the response.
```

**Prompt for "DBpedia-14" dataset**

```
You will have the role of a researcher that performs content
analysis tasks.

I need you to code a series of variables based on movie
reviews. Next, I will describe you the variables.

Variable "sentiment" which represents the overall sentiment or
emotional tone expressed in the movie review. Answer with one
of the following values: Negative, Positive

All the variables are defined. This is the content to analyze
(between triple quotes):

"""{text}"""

Return the response in JSON format with a dictionary with a
key/value pair for each variable. Do not include any other
content in the response.
```

**Prompt 1 for the "SciCite" dataset (providing description of each citation intent category)**

```
You will have the role of a researcher that performs content
analysis tasks.

I need you to code a series of variables based on scientific
citation contexts. Next, I will describe you the variables.

Variable "citation_intent" which represents the author's
intent or purpose when citing a scientific paper in their
research work. Answer with one of the following values:
background, method, result

Classification rules:
- Assign 'background' if the citation intent is "States,
mentions, or points to the background information giving more
context about a problem, concept, approach, topic, or
importance of the problem in the field."
- Assign 'method' if the citation intent is "Making use of a
method, tool, approach or dataset."
- Assign 'result' if the citation intent is "Reports or
comparing paper's results/findings with the results/findings
of other work."
- Return only a single value in your response.

All the variables are defined. This is the content to analyze
(between triple quotes):

"""{text}"""

Return the response in JSON format with a dictionary with a
key/value pair for each variable. Do not include any other
content in the response.
```

**Prompt 2 for the "SciCite" dataset (only names of categories)**

```
You will have the role of a researcher that performs content
analysis tasks.

I need you to code a series of variables based on scientific
citation contexts. Next, I will describe you the variables.

Variable "citation_intent" which represents the author's
intent or purpose when citing one or more scientific papers in
their research work. Answer with one of the following values:
background, method, result

When multiple cites appear together, answer with a single
value for all of them.

All the variables are defined. This is the content to analyze
 (between triple quotes):

"""{text}"""

Return the response in JSON format with a dictionary with a
key/value pair for each variable. Do not include any other
content in the response.
```